# On the Optimal Solution of Weighted Nuclear Norm Minimization

Qi Xie[a], Deyu Meng[a], Shuhang Gu[b], Lei Zhang[b], Wangmeng Zuo[c],
Xiangchu Feng[d] and Zongben Xu[a]

[a]School of Mathematics and Statistics, Xi'an Jiaotong University, Xi'an, China
[b]Dept. of Computing, The Hong Kong Polytechnic University, Hong Kong
[c]Dept. of Computer Science, Harbin Institute of Technology, Harbin, China
[d]Dept. of Applied Mathematics, Xidian University, Xi'an, China

**Abstract**

In recent years, the nuclear norm minimization (NNM) problem has been attracting much attention in computer vision and machine learning. The NNM problem is capitalized on its convexity and it can be solved efficiently. The standard nuclear norm regularizes all singular values equally, which is however not flexible enough to fit real scenarios. Weighted nuclear norm minimization (WNNM) is a natural extension and generalization of NNM. By assigning properly different weights to different singular values, WNNM can lead to state-of-the-art results in applications such as image denoising [3]. Nevertheless, so far the global optimal solution of WNNM problem is not completely solved yet due to its non-convexity in general cases. In this article, we study the theoretical properties of WNNM and prove that WNNM can be equivalently transformed into a quadratic programming problem with linear constraints. This implies that WNNM is equivalent to a convex problem and its global optimum can be readily achieved by off-the-shelf convex optimization solvers. We further show that when the weights are non-descending, the globally optimal solution of WNNM can be obtained in closed-form.

**1. Introduction**

As a classical technique for low rank matrix approximation, recently the nuclear norm minimization (NNM) [1-2] has been attracting much attention in computer vision and machine learning research. The standard nuclear norm of a matrix $X \in \mathbb{R}^{m \times n}$ is defined as the sum of all its singular values, i.e., $\|X\|_* = \sum_i |\sigma_i(X)|$, where $\sigma_i(X)$ is the $i$-th singular value of $X$. Nuclear norm is the tightest convex relaxation of the rank penalty of a matrix. Let $Y \in \mathbb{R}^{m \times n}$ be the given data matrix. The standard NNM problem aims to find an approximation matrix $X$ of $Y$ by minimizing the following energy function:

$$\min_{X} \|Y - X\|_F^2 + \lambda \|X\|_*, \tag{1}$$

where $\lambda$ is a positive regularization parameter. It has been shown that the above NNM problem has a closed-form solution [2]:

$$X^* = U \mathcal{S}_\lambda(\Sigma) V^T,$$

where $Y = U\Sigma V^T$ is the SVD of $Y$, and $\mathcal{S}_\lambda(\Sigma) = \max(0, \Sigma - \lambda/2)$.

Albeit easy to solve, the NNM model has some limitations. The nuclear norm treats all the singular values equally, and it ignores the prior knowledge we often have on the matrix singular values. For example, in many vision applications, the larger singular values of the data matrix are usually more important than the smaller ones since they represent the main components of the data. Intuitively, we should assign different weights to different singular values to make NNM more flexible to fit real scenarios.

To improve the flexibility of NNM, researchers have proposed the weighted nuclear norm minimization

(WNNM) [3-4] problem. The weighted unclear norm of a matrix $X$ is defined as:
$$\|X\|_{w,*} = \sum_i |w_i \sigma_i(X)|, \qquad (2)$$
where $\sigma_1(X) \geq \sigma_2(X) \geq \cdots \geq \sigma_n(X)$, $w = [w_1, w_2, \ldots, w_n]$ and $w_i \geq 0$ is the weight assigned to $\sigma_i(X)$. Weighted nuclear norm is not a real norm since it does not always satisfy the triangle inequality. The WNNM problem is then formulated as [3]:
$$\min_X \|Y - X\|_F^2 + \|X\|_{w,*}. \qquad (3)$$

The WNNM problem, however, is more difficult to solve than the NNM problem due to its non-convexity in general cases of weights. Gu et al. [3] and Yang et al. [4] have independently discussed the solution of WNNM and successfully applied WNNM to image denoising and 3D reconstruction. Nonetheless, the globally optimal solution of WNNM is not completely solved yet.

In this article, we aim to study and present the optimal solution of WNNM in Eq. (3). We prove that the WNNM problem can be equivalently transformed into a convex programming problem, and its globally optimal solution can be readily achieved by employing off-the-shelf quadratic programming techniques. We further show that in a specific but very useful case, i.e., the weights are in a non-descending order, the global minimum can be analytically obtained in closed form.

## 2. Low-Rank Minimization with Weighted Nuclear Norm

We first give the following Lemma 1 [3], which builds the important relationship between the nuclear norm and the trace of a matrix.

**Lemma 1[3].** *For any $A \in \mathbb{R}^{m \times n}$ and a diagonal non-negative matrix $W \in \mathbb{R}^{m \times n}$ with non-ascending ordered diagonal elements, let $A = X \Phi Y^T$ be the SVD of $A$, we have*
$$\sum_i \sigma_i(A) \sigma_i(W) = \max_{U^T U = I, V^T V = I} tr[W U^T A V],$$
*where $I$ is the identity matrix, $\sigma_i(A)$ and $\sigma_i(W)$ are the i-th singular values of matrices $A$ and $W$, respectively. When $U = X$ and $V = Y$, $tr[W U^T A V]$ reaches its maximum value.*

Based on the result of Lemma 1, we can have the following theorem, which implies an equivalent convex transform of the original non-convex WNNM problem in Eq. (3).

**Theorem 1.** *For any $Y \in \mathbb{R}^{m \times n}$, without loss of generality we assume that $m \geq n$, and let $Y = U \Sigma V^T$ be the SVD of $Y$, where $\Sigma = \begin{pmatrix} diag(\sigma_1, \sigma_2, \ldots, \sigma_n) \\ 0 \end{pmatrix}$. The solution of the WNNM problem in Eq. (3) can be expressed as $X^* = U D V^T$, where $D = \begin{pmatrix} diag(d_1, d_2, \ldots, d_n) \\ 0 \end{pmatrix}$ is a diagonal non-negative matrix and $(d_1, d_2, \ldots, d_n)$ is the solution of the following convex optimization problem:*
$$\min_{d_1, d_2, \ldots, d_n} \sum_{i=1}^n (d_i - \sigma_i)^2 + w_i d_i, \text{ s.t. } d_1 \geq d_2 \geq \ldots \geq d_n \geq 0. \qquad (4)$$

***Proof.*** For any $X \in \mathbb{R}^{m \times n}$, its SVD can be expressed as $X = \bar{U} D \bar{V}^T$, where $\bar{U}$ and $\bar{V}$ are unitary matrices, and $D = \begin{pmatrix} diag(d_1, d_2, \ldots, d_n) \\ 0 \end{pmatrix}$ with $d_1 \geq d_2 \geq \ldots \geq d_n \geq 0$. Then we have

$$\min_{X} \|Y - X\|_F^2 + \|X\|_{w,*} \Leftrightarrow$$

$$\min_{\bar{U},\bar{V},D} \|Y - \bar{U}D\bar{V}^T\|_F^2 + \|\bar{U}D\bar{V}^T\|_{w,*} \quad s.t. \quad d_1 \geq d_2 \geq \ldots \geq d_n \geq 0 \Leftrightarrow$$

$$\min_{\bar{U},\bar{V},D} \|Y\|_F^2 - 2tr(Y\bar{U}^T D\bar{V}) + \|D\|_F^2 + \|D\|_{w,*} \quad s.t. \quad d_1 \geq d_2 \geq \ldots \geq d_n \geq 0 \Leftrightarrow$$

$$\min_{D} \left( \|D\|_F^2 + \|D\|_{w,*} - 2 \max_{\bar{U}^T\bar{U}=I, \bar{V}^T\bar{V}=I} tr(Y\bar{U}^T D\bar{V}) \right) \quad s.t. \quad d_1 \geq d_2 \geq \ldots \geq d_n \geq 0.$$

According to Lemma 1, we have

$$\max_{\bar{U}^T\bar{U}=I, \bar{V}^T\bar{V}=I} tr(Y\bar{U}^T D\bar{V}) = \sum_i \sigma_i d_i,$$

and the optimal solution is obtained at $\bar{U} = U$ and $\bar{V} = V$. We then have

$$\min_{X} \|Y - X\|_F^2 + \|X\|_{w,*} \Leftrightarrow$$

$$\min_{D} \left( \|D\|_F^2 + \|D\|_{w,*} - 2\sum_{i=1}^{n} d_i \sigma_i \right) \quad s.t. \quad d_1 \geq d_2 \geq \ldots \geq d_n \geq 0 \Leftrightarrow$$

$$\min_{D} \left( \sum_{i=1}^{n} d_i^2 + w_i d_i - 2 d_i \sigma_i \right) \quad s.t. \quad d_1 \geq d_2 \geq \ldots \geq d_n \geq 0 \Leftrightarrow$$

$$\min_{D} \left( \sum_{i=1}^{n} (d_i - \sigma_i)^2 + w_i d_i \right) \quad s.t. \quad d_1 \geq d_2 \geq \ldots \geq d_n \geq 0.$$

From the above derivation, we can see that the optimal solution of the WNNM problem in Eq. (3) is

$$X^* = UDV^T, \tag{5}$$

where $D$ is the optimum of the constrained optimization problem in Eq. (4). The proof is then completed. ∎

Theorem 1 shows that the WNNM problem can be equivalently transformed into the problem in Eq. (4). It is interesting to see that Eq. (4) is a convex problem. This means that the original difficult non-convex problem is equivalent to a convex problem which is much easier to solve. Furthermore, in a specific yet very useful case, i.e., the weights $w_{i=1,2,\ldots n}$ are in a non-descending order, it can be shown that the global optimum of Eq. (4) has a closed form. We have the following corollary.

**Corollary 1.** *If the weights satisfy $0 \leq w_1 \leq w_2 \leq \ldots \leq w_n$, the globally optimal solution of Eq. (4) is $D^* = \begin{pmatrix} diag(\bar{d}_1, \bar{d}_2, \ldots, \bar{d}_n) \\ 0 \end{pmatrix}$, where $\bar{d}_i = \max(\sigma_i - \frac{w_i}{2}, 0)$.*

***Proof.*** If we ignore the constraints of the problem in Eq. (4), we can obtain the following unconstrained problem:

$$\min_{d_i \geq 0} (d_i - \sigma_i)^2 + w_i d_i. \tag{6}$$

Since $d_i \geq 0$, $i = 1,2,\ldots,n$, the above problem is equivalent to the following problem:

$$\min_{d_i \geq 0} \left( d_i - \left( \sigma_i - \frac{w_i}{2} \right) \right)^2.$$

Then it is easy to see that the globally optimal solution of the problem in Eq. (6) is

$$\bar{d}_i = \max\left( \sigma_i - \frac{w_i}{2}, 0 \right), i = 1,2,\ldots,n.$$

Since $\sigma_1 \geq \sigma_2 \geq \ldots \geq \sigma_n$ and $w_1 \leq w_2 \leq \ldots \leq w_n$, we have
$$\bar{d}_1 \geq \bar{d}_2 \geq \cdots \geq \bar{d}_n \geq 0.$$
Thus, $\bar{d}_{i=1,2,\ldots,n}$ satisfy the constraint of Eq. (4), and this implies that they are the solution of the original constrained problem in Eq. (4). The proof is then completed. ∎

The conclusion in Corollary 1 is very useful in some real scenarios. For instance, Gu et al. [3] have shown that by assigning smaller weights to the larger singular values, WNNM leads to state-of-the-art image denoising results. Actually, combining Theorem 1 and Corollary 1, we can readily get the globally optimal solution $X^*$ in the following analytical form (when $0 \leq w_1 \leq w_2 \leq \ldots \leq w_n$):
$$X^* = UD^*V^T, \tag{7}$$
where
$$D^* = \begin{pmatrix} diag(\bar{d}_1, \bar{d}_2, \ldots, \bar{d}_n) \\ 0 \end{pmatrix}, \ \bar{d}_i = \max(\sigma_i - \frac{w_i}{2}, 0), \tag{8}$$
and $Y = U\Sigma V^T$ is SVD of $Y$. This reveals the underlying reason of the effectiveness of the WNNM denoising method in [3].

Finally, we summarize how to calculate the global optimum of the WNNM problem as follows:
- *If the weights satisfy $0 \leq w_1 \leq w_2 \leq \ldots \leq w_n$, the globally optimal solution of the WNNM problem with input matrix $Y$ can be expressed as $X^* = UD^*V^T$, where $U\Sigma V^T$ is the SVD of $Y$ and $D^* = \begin{pmatrix} diag(\bar{d}_1, \bar{d}_2, \ldots, \bar{d}_n) \\ 0 \end{pmatrix}$ has a closed form solution as presented in Corollary 1.*
- *In the general case, i.e., the weights are in an arbitrary order, the global optimum of the WNNM problem with input matrix $Y$ can be expressed as $X^* = UDV^T$, where $U\Sigma V^T$ is the SVD of $Y$ and $D = \begin{pmatrix} diag(\bar{d}_1, \bar{d}_2, \ldots, \bar{d}_n) \\ 0 \end{pmatrix}$ can be calculated by quadratic programming with linear constraints:*
$$\min_{d_1, d_2, \ldots, d_n} \sum_{i=1}^n d_i^2 + (w_i - 2\sigma_i)d_i \ \ \text{s.t.} \ \ d_1 - d_2 \geq 0, d_2 - d_3 \geq 0, \ldots, d_{n-1} - d_n \geq 0, d_n \geq 0.$$

*It should be noted that the above quadratic programming problem can be accurately and efficiently solved by many off-the-shelf optimization toolkits [5, 6, 7].*

## 3. Conclusion

In this article, we studied the theoretical properties of the WNNM problem. When the weights are in an arbitrary order, we showed that WNNM can be equivalently transformed into a quadratic programming problem which is easy to solve by off-the-shelf toolkits. When the weights are in a non-descending order, interestingly, we proved that the globally optimal solution of WNNM can be obtained in closed form. Our findings reveal that although the WNNM problem is non-convex, its global optimum can still be obtained. It is thus expected that the WNNM model will have more successful applications in computer vision and machine learning.